%% file: acl_latex.tex
\documentclass[11pt]{article}

\usepackage[preprint]{acl}

\usepackage{times}
\usepackage{latexsym}

\usepackage[T1]{fontenc}

\usepackage[utf8]{inputenc}

\usepackage{microtype}

\usepackage{inconsolata}
\usepackage{todonotes}
\usepackage{url}
\usepackage{booktabs}
\usepackage{multirow}
\usepackage{threeparttable}

\usepackage{graphicx}

\definecolor{BetterBlue}{RGB}{0,92,175}
\definecolor{WorseRed}{RGB}{180,35,24}

\title{Enhancing Mental Health Counseling Support in Bangladesh using 
Culturally-Grounded Knowledge
}

\author{Md Arid Hasan$^1$, Azhagu Meena SP$^1$, Aditya Khan$^1$, \\ 
\textbf{Abu Md Akteruzzaman Bhuiyan$^2$, Helal Uddin Ahmed$^3$, Joysree Debi$^4$,} \\ \textbf{Farig Sadeque$^5$, Annie En-Shiun Lee$^1$ and Syed Ishtiaque Ahmed$^1$} \\
$^1$University of Toronto, Canada \\
$^2$mPower Social Enterprises Ltd, Bangladesh, $^3$National Institute of Mental Health, Bangladesh \\
$^4$Sajida Foundation, Bangladesh, $^5$BRAC University, Bangladesh \\
{\tt \{arid, ishtiaque\}@cs.toronto.edu}
}

\begin{document}
\maketitle
\begin{abstract}
Large language models (LLMs) show promise in generating supportive responses for mental health and counseling applications. However, their responses often lack cultural sensitivity, contextual grounding, and clinically appropriate guidance. 
This work addresses the gap of how to systematically incorporate domain-specific, clinically validated knowledge into LLMs to improve counseling quality. We utilize and compare two approaches, retrieval-augmented generation (RAG) and a knowledge graph (KG)–based method, designed to support para-counselors. 
Our KG is constructed manually and clinically validated, capturing causal relationships between stressors, interventions, and outcomes, with contributions from multidisciplinary people. 
We evaluated multiple LLMs in both settings using BERTScore F1 and SBERT cosine similarity, as well as human evaluation across five metrics, 
which is designed to directly measure the effectiveness of counseling beyond similarity at the surface level. 
The results show that KG-based approaches consistently improve contextual relevance, clinical appropriateness, and practical usability compared to RAG alone, demonstrating that structured, expert-validated knowledge plays a critical role in addressing LLMs limitations in counseling tasks. 


\end{abstract}

\input{sections/introduction}

\input{sections/related_work}

\input{sections/methodology}

\input{sections/results}

\input{sections/conclusion}

\section{Limitations}
Despite the promising results, this study has several limitations. First, the dataset used in this work is relatively small, consisting of a limited number of annotated counseling cases. Although the narratives were carefully validated by domain experts, the dataset may not fully capture the diversity of mental health experiences, cultural contexts, and counseling scenarios encountered in real-world settings. Second, while the knowledge graph was manually curated by a group of experts to ensure clinical validity, the scope of the graph is constrained by the available cases and domain coverage. As a result, some complex relationships between stressors, contextual factors, and interventions may not yet be fully represented.

Although both automatic and human evaluations were conducted, human evaluation inherently involves subjective judgment, and the number of evaluators and evaluation rounds may limit the generalizability of the results. Finally, the system is designed to support counseling guidance rather than replace professional clinical judgment. Therefore, responses generated by the model should be interpreted as supportive suggestions rather than definitive clinical recommendations.

\section*{Ethical Consideration}
Our research examines the feasibility of providing mental health support to people in lower-income communities through para-counselors. Much of the existing work on culturally sensitive LLMs focuses on direct user-facing applications, where marginalized individuals interact with chatbots for emotional support \cite{Ma2024-hw, Aleem2024-me}. Given the ethical and clinical risks involved, scholars have explicitly warned against using LLMs as substitutes for licensed therapists \cite{moore2025expressing}. Concerns include the absence of clinical accountability, potential for harmful misinterpretation, and lack of sustained relational understanding.

Informed by this body of work, we are studying the use of LLMs by para-counselors rather than directly by service users. Our work represents a potentially promising and ethically moderated application. Here, LLMs would serve as knowledge-support systems to augment para-counselors’ decision-making capacities, while ultimate responsibility for interpretation and action remains with the human provider.

\bibliography{custom}

\appendix

\input{sections/appendix}

\end{document}

%% file: sections/introduction.tex
\section{Introduction}
\label{sec:introduction}
Mental health systems in developing countries such as Bangladesh remain severely under-resourced, with significant constraints in both financial investment and trained mental health personnel \cite{rahaman2025mental, jahan2024ngos, harrison2019strategic}. In response to these resource constraints, Bangladesh, like many developing nations, has increasingly relied on task-shifting approaches, in which non-specialists, called para-counselors, with minimal training (ranging from several weeks to one month) in basic counseling skills and mental health first aid, deliver frontline mental health support \cite{mariam2023sisters}.

\begin{figure}[t]
\centering
\includegraphics[width=0.7\linewidth]{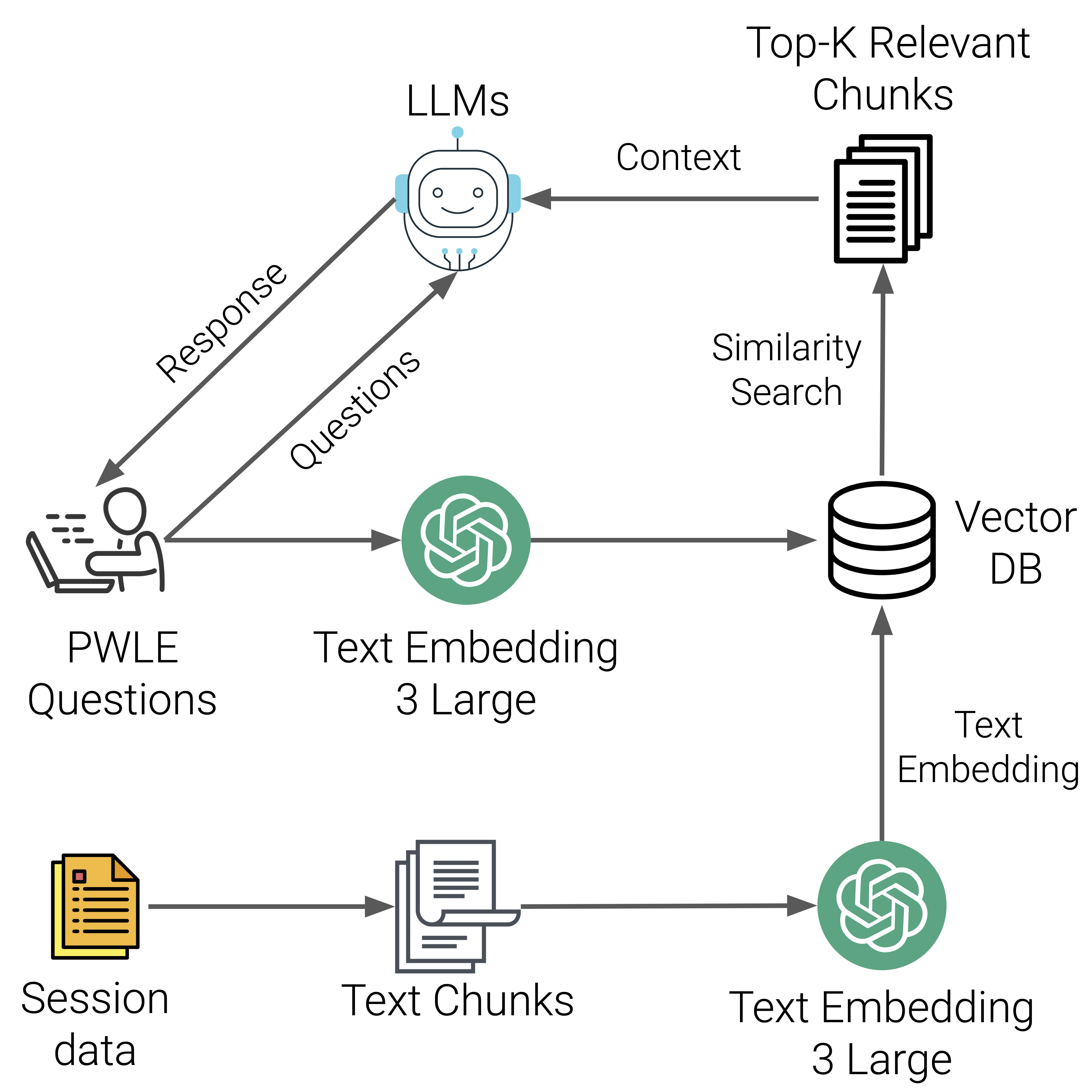}
\caption{Overview of the methodology incorporating RAG-based experiments.}
\vspace{-0.6cm}
\label{fig:methodology}
\end{figure}

In the context of the present study, para-counselors occupy an intermediary position between community health workers and mental health specialists. Community health workers identify and refer individuals presenting with psychological distress. Para-counselors then conduct a structured series of sessions, typically four to six, depending on the severity and vulnerability of the case. If they identify that a community member needs professional support, they may refer the individual to a psychologist, psychiatrist, or a tele-mental health service \cite{mariam2023sisters}. 

Despite their pivotal role, para-counselors in Bangladesh, like lay counselors and frontline health workers in other developing nations, often operate with minimal supervision, limited refresher training, and limited access to real-time clinical guidance for providing culturally sensitive care \cite{singla2022culturally}. Addressing these systemic limitations is essential to strengthening community mental health care and ensuring that task-shifting models translate into safe, effective, and equitable service delivery. To address these concerns, recent research, though it has not focused on para-counselors particularly, has explored the potential of using LLMs to support frontline health workers in meeting knowledge and decision-support needs \cite{ramjee2025ashabot, moosa2025evaluating, lima2025spotlighting, rutunda2025good, mastel2025mbaza, al2023transforming}.

However, while this body of work acknowledges the possible advantages of LLM-powered tools, it also underscores important ethical and contextual risks \cite{ramjee2025ashabot, lima2025spotlighting}. Scholars suggest that for an effective collaboration between LLM-based systems and healthcare workers to meet frontline health workers' knowledge needs, the models must show a deep understanding of the sociocultural contexts in which they are deployed \cite{ramjee2025ashabot, lima2025spotlighting}. Without such value alignment, LLMs might risk reproducing Western norms and assumptions, potentially imposing culturally incongruent interpretations of distress or overlooking structural issues such as patriarchy and economic precarity \cite{song2025typing, Aleem2024-me}. 

Hence, we propose a generative AI support system that, in two non-diagnostic ways, meets para-counselors' knowledge-support needs. First, the system acts as a symptom-reflection and messaging assistant, helping para-counselors draft context-sensitive responses based on session summaries. Second, it functions as a query-based educational tool, providing real-time guidance on therapeutic strategies and locally grounded idioms of care. To facilitate the effective development of such a system, we investigate two key research questions: \textit{(i)} to what extent do LLMs reflect cultural understandings of mental health when used by para-counselors supporting low-income communities in Bangladesh? and \textit{(ii)} how can LLMs deployed to support lower-income communities in Bangladesh be designed to be more culturally sensitive?

To address these research questions, we first constructed a dataset from initial counseling sessions conducted by para-counselors with lower-income community members and used it within a retrieval-augment generation (RAG) framework to provide contextual grounding for LLMs, given their known limitations in cultural sensitivity \cite{liu2026tailored}. We then developed a culturally grounded knowledge graph (KG) through a co-design process involving para-counselors, persons with lived experience (PWLEs), public health researchers (PHRs), psychologists, and psychiatrists to experiment with the KG-based framework. KG captures causal relationships among triggers, symptoms, and coping strategies (e.g., economic hardship $\rightarrow$ sleep loss $\rightarrow$ perceived spiritual affliction), as well as local idioms of distress, to better ground LLM responses. 

Our findings highlight the importance of socio-technical approaches for improving the cultural sensitivity of the LLM-based knowledge-support system of para-counselors. We found that the LLM with structured knowledge yields significantly better quality and practical relevance than RAG-based models that rely on unstructured data. Further, quantitative evaluations across four language models, assessing fluency, relevance, and cultural alignment, complemented with human evaluations from para-counselors, PHRs, PWLE, psychologists, and psychiatrists, indicate that effective counseling support depends not only on semantic similarity to reference responses but also on clear language, accurate interpretation of concerns, appropriate guidance, and contextual awareness.

%% file: sections/related_work.tex
\vspace{-0.1cm}
\section{Related Work}
\label{sec:related_work}

\vspace{-0.3em}
\subsection{LLMs and Mental Health}

LLMs are increasingly being explored as tools to expand access to mental health services, particularly in resource-constrained settings where trained professionals and infrastructure are limited. Some of the potential research directions in this area includes predicting mental health states from social media data \cite{zhai2024chinese, juhng2023discourse}, detecting mental health disorders \cite{song2025does}, assessing mental health risk \cite{zheng-etal-2025-promind}, conducting mental health analysis \cite{yang2023towards}, automating specific therapeutic techniques, such as cognitive behavioral therapy or motivational interviewing \cite{Sharma2023CognitiveRO, mahmood2025fully}, providing personalized emotional support \cite{cheng2023pal}, and generating synthetic or simulating realistic patients, enabling trainee psychologists or counselors to practice therapeutic interactions without requiring human participants \cite{lozoya2024generating,wang2025annaagent}. 

To address different research objectives, scholars have explored both technical and socio-technical approaches \cite{lozoya2024generating, zheng-etal-2025-promind, cho2023integrative}. Technical approaches include prompting strategies for improving the quality of synthetic data \cite{lozoya2024generating}, fine-tuning models on domain-specific datasets curated by experts, and incorporating objective behavioral data alongside subjective data to improve predictive performance \cite{zheng-etal-2025-promind}, and reasoning distillation to improve the performance of small language models used for mental health disorder detection and rationale generation \cite{song2025does}. Socio-technical approaches emphasize collaboration and stakeholder involvement in the development of LLM-based mental health systems. This includes facilitating multidisciplinary collaboration between mental health professionals and NLP researchers \cite{cho2023integrative}. However, not many studies have investigated how the inclusion of key stakeholders, such as para-counselors or individuals with lived experience of mental health challenges, can shape the design and evaluation of these systems.

\vspace{-0.1cm}
\subsection{LLMs for Para-Counselors}
Most works that use LLMs for mental health support primarily focus on improving the general capacity of LLMs for various downstream mental health tasks \cite{sabour2024emobench}, supporting trained psychologists \cite{lozoya2024generating,wang2025annaagent} or users who could directly use these systems \cite{mahmood2025fully}. Comparatively little attention has been given to how frontline mental health workers, such as para-counselors who often support people with little or no formal education, unstable working conditions, or unemployment, might benefit from such systems. 

Notably, most existing studies are conducted largely in Western or Chinese contexts, with relatively limited design and evaluation in developing regions \cite{cho2023integrative}. Research examining LLM-based mental health systems across diverse cultural settings has highlighted important cultural and linguistic blind spots \cite{Aleem2024-me, Ma2024-hw}. As these models are typically trained and evaluated using data grounded in Western psychological frameworks and cultural contexts, they may fail to capture culturally specific idioms of distress, local healing practices, and socially embedded understandings of well-being \cite{song2025typing}. As a result, such systems risk universalizing particular conceptions of mental illness, diagnosis, and treatment while overlooking alternative epistemologies and community-based perspectives on mental health \cite{flore2023artefacts}. These limitations suggest the need for more culturally grounded design approaches.

%% file: sections/methodology.tex
\vspace{-0.2cm}
\section{Methodology}
\label{sec:methodology}
\vspace{-0.1cm}


Figure~\ref{fig:methodology} illustrates the overall architecture of our methodology. Our approach is motivated by the limitation that LLMs, while well known for their superior performance on various NLP downstream tasks~\cite{minaee2024large}, often lack access to domain-specific, culturally grounded, and clinically validated knowledge, which is critical for effective counseling support. To address this, we first construct a dataset of counseling sessions that captures real-world stressors, contextual factors, interventions, and outcomes—information that is typically underrepresented in general-purpose LLM training data \cite{xu2024mental}. This dataset is further structured into a knowledge graph (KG) to explicitly model causal relationships among problems, interventions, and outcomes, enabling more interpretable and clinically grounded reasoning.

To leverage both unstructured and structured knowledge, we integrate retrieval-augmented generation (RAG) and KG-based experiments. Session data are embedded and stored in a vector database, allowing the system to retrieve the top-$K$ semantically relevant cases when a PWLE submits a query. These retrieved contexts provide situational basis for RAG-based approach, while the KG contributes structured, expert-validated relationships that guide reasoning about appropriate interventions. By using KG, the LLM is able to generate responses that are more contextually relevant, clinically appropriate, and aligned with real-world counseling practices.

\vspace{-0.5em}
\subsection{Dataset}
\label{sec:dataset}
This section provides a detailed description of the dataset development process, including data collection, annotation, and KG construction.

\begin{figure*}[t]
\centering
\includegraphics[width=\linewidth]{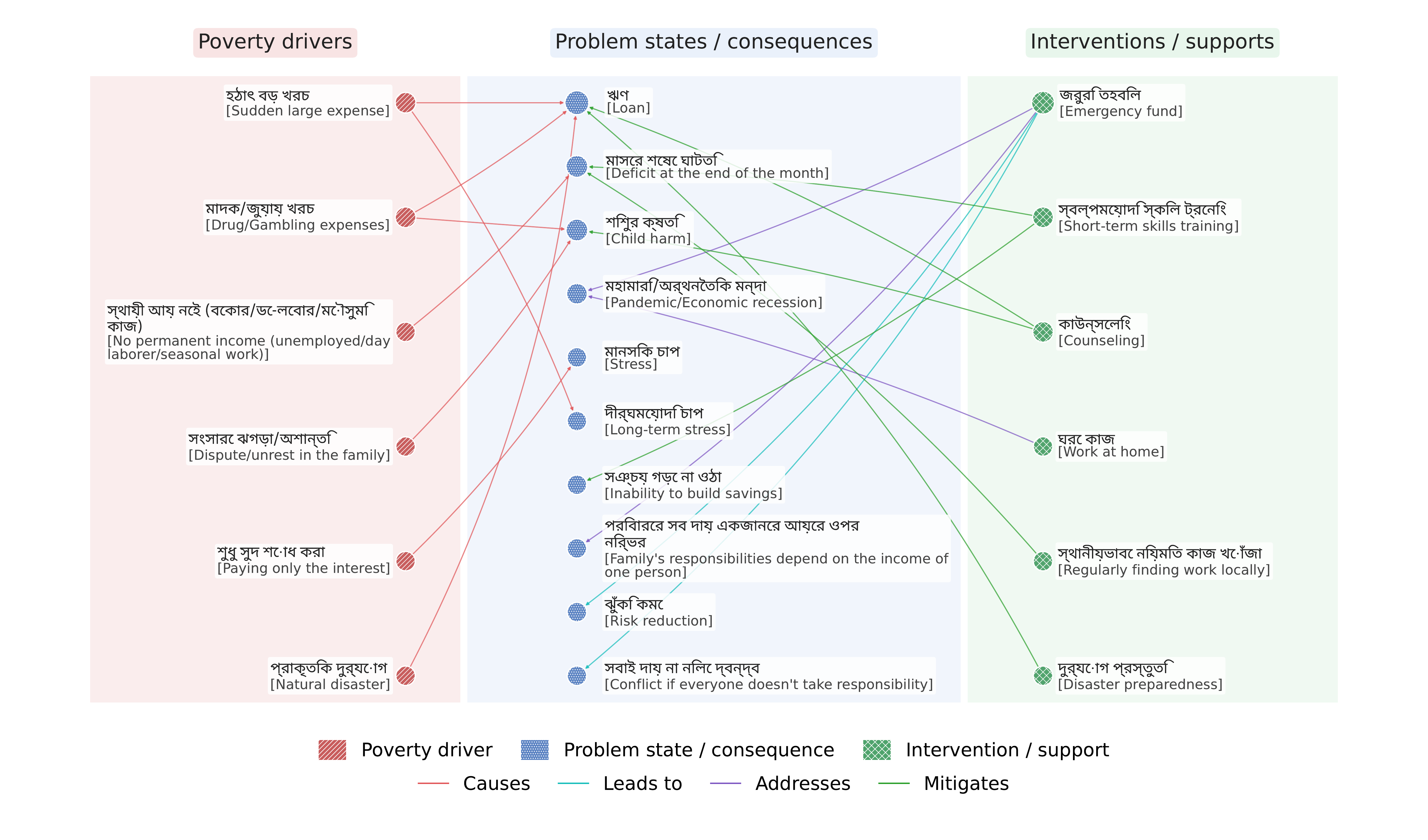}
\vspace{-0.8cm}
\caption{Illustrative subgraph of the knowledge graph. 
Nodes are grouped into three broad components: poverty drivers, problem states or consequences, and interventions or supports. Directed edges denote four relation types: 
\textit{Causes}, \textit{Leads to}, \textit{Addresses}, and \textit{Mitigates}. For readability, this figure shows only a curated, representative subset of the most salient nodes and relations used in the system, rather than the full knowledge graph. The complete graph is substantially larger and contains additional nodes, edges, and relation instances not displayed here.}
\vspace{-0.6cm}
\label{fig:poverty-kg}
\end{figure*}

\vspace{-0.3em}
\subsubsection{Data Collection}

Our dataset is constructed from a collection of structured counseling case records written in Bangla. Each record documents the demographic characteristics of an individual (e.g., age, sex, occupation, marital status), reported causes of distress, and narrative notes prepared for counseling sessions. The sessions were conducted by para-counselors as part of Sajida Foundation's Community Mental Health Initiative. The cases reflect real-world mental health intake scenarios and thus provide naturalistic descriptions of emotional states, stressors, and behavioral symptoms. Each record consists of 2-6 sessions among the participants and a para-counselor. We also provide a strict guideline to the para-counselors on conducting sessions, which we have discussed in Appendix~\ref{appn:guideline_session}.

Across all sessions, para-counselors record the participant’s problems, contextual factors (such as family, economic, or social influences), emotional responses, and the interventions or guidance provided. This structured approach ensures that the collected data captures both the counseling process and the evolution of the participant’s situation over time. As a result, we obtained $402$ participant records from Sajida Foundation's Community Mental Health Initiative. Participants vary in age, occupation, marital status, and socioeconomic background, reflecting diverse life circumstances and sources of psychological distress. The records include both short categorical fields (e.g., gender, employment status) and longer narrative descriptions capturing emotional states, perceived causes of stress, and recommended interventions.

\subsubsection{Manual Annotation}
Following the annotation guidelines in Appendix~\ref{sec:annotation_guideline}, the manual annotation process was conducted in two structured steps to ensure narrative quality, privacy compliance, and clinical validity. Our pool of manual annotators consists of 6 individuals, including one person with lived experience (PWLE), two public health researchers (PHRs), two Psychologists, and one Psychiatrist. In the first step, annotators 
removed direct identifiers (e.g., names, workplaces, locations) and generalized specific dates using relative session references. This step was carried out collaboratively by one PWLE and two PHRs, who independently reviewed cases and documented suggested corrections before reconciling differences through discussion.

In second step, two licensed psychologists independently evaluated whether the narrative contained sufficient information to establish a meaningful cause and effect relationship, whether protective and risk factors were correctly categorized, and whether the intervention type was clinically appropriate given the participant’s demographics and context. The annotation task was evenly distributed between psychologists. If a psychologist raised concerns about certain cases, particularly regarding the appropriateness of the clinical interpretation, a discussion was arranged among the psychologists and the psychiatrist to reach a final adjudication and resolve discrepancies.  
This multi-disciplinary annotation framework ensured both methodological rigor and clinical reliability in the final labeled dataset. As a result of the manual annotation process, we obtained a final dataset of 69 individual annotated cases. We provide a detailed analysis of the data in Appendix~\ref{appn:data_analysis}.

\begin{table}[!ht]
\centering
\scalebox{0.8}{
\setlength{\tabcolsep}{4pt}
\begin{tabular}{lrp{4.5cm}}
\toprule
\textbf{Relationship} & \textbf{Count} & \textbf{Connection Logic} \\ \midrule
MITIGATES & 368 & Intervention $\rightarrow$ Effect \\
LEADS\_TO & 184 & Intervention $\rightarrow$ Outcome \\
CAUSES & 92 & Cause $\rightarrow$ Effect \\
ADDRESSES & 92 & Intervention $\rightarrow$ Cause \\
BELONGS\_TO & 23 & Node $\rightarrow$ Category \\
COMPLEMENTS & 20 & Intervention $\rightarrow$ Intervention \\
RELATED\_TO & 20 & Node $\rightarrow$ Node \\
EXACERBATES & 14 & Cause $\rightarrow$ Cause \\
REQUIRES & 9 & Intervention $\rightarrow$ Intervention \\
\bottomrule
\end{tabular}
}
\caption{Relationship distribution of the knowledge graph.}
\vspace{-0.4cm}
\label{tab:kg_stats}
\end{table}

\subsubsection{Knowledge Graph Construction}
Following manual annotation, the annotated cases were transformed into a structured knowledge graph (KG) to represent relationships among causes, interventions, effects, and outcomes in counseling narratives. The nodes and relationships of KG were selected in a co-design workshop in the presence of PWLE, para-counselors, psychologists, and psychiatrists, ensuring that the relationships reflect clinically meaningful reasoning and domain expertise. Once the KG was built, it was 
migrated into a graph database using relational queries, converting static annotations into a dynamic and traversable relational structure. This process enabled the modeling of causal and intervention pathways within mental health support scenarios.

The resulting knowledge graph contains 308 nodes and 642 relationships across nine relationship types. The structure is dense and multi-dimensional, capturing complex connections between causes, protective factors, interventions, and outcomes in a form that can be systematically explored and analyzed. The nodes represent domain entities such as causes, effects, interventions, outcomes, and conceptual categories, while the edges encode the semantic relationships between them. We provide the detailed statistics of KG in Table~\ref{tab:kg_stats}. Moreover, we provide an overview of the KG with a subset of data in Figure~\ref{fig:poverty-kg}.

\subsection{Models}
Recent advances in large language models (LLMs) have attracted substantial research interest, particularly in assessing their effectiveness within healthcare settings and across diverse downstream medical applications \cite{maity2025large}. 
To conduct our experiments, we evaluated models from multiple families—including 
Gemini, Gemma, Llama, DeepSeek, and GPT—to enable a broad comparative analysis. Specifically, we tested several variants from each family, covering both lightweight and large-scale models, to examine performance differences across architectures and model sizes. This allowed us to systematically analyze how model choice affects RAG- and KG-based experiments, and to identify which configurations yield more accurate, contextually relevant, and clinically appropriate counseling responses.

\subsection{Experimental Settings}
We experimented with two different settings: RAG-based and KG-based approaches. In RAG-based setting, relevant information was retrieved from the annotated dataset and incorporated into the model’s input context to support response generation. In the KG-based setting, the model leveraged structured knowledge from the constructed knowledge graph, enabling reasoning over explicit relationships between causes, interventions, effects, and outcomes. This comparison allows us to evaluate how unstructured retrieval and structured knowledge representations influence the quality and relevance of generated responses.

\subsubsection{RAG-based Experiment Setting}
For RAG-based experiments, we employed a dense retrieval pipeline using the Text Embedding 3 Large embedding model from OpenAI to encode the dataset for semantic search. The retrieval process prioritized dense embeddings with a dense-to-sparse weighting of 1, meaning retrieval relied entirely on dense vector similarity. To construct the context for generation, the system retrieved up to three document citations, with each snippet limited to 500 words and an overlap ratio of 2 to preserve contextual continuity across chunks.

For response generation, we configured the language model with a maximum output length of 7,960 tokens to accommodate detailed reasoning and explanations. The temperature was set to 0.2, encouraging more deterministic and consistent responses while minimizing randomness during generation. This configuration was designed to balance retrieval relevance, contextual coverage, and controlled response generation within the RAG framework. We adopted the same configuration and hyperparameter settings across all models in the RAG-based experiment to ensure a fair and consistent comparison.

\subsubsection{KG-based Experiment Setting}
For KG-based experiments, we integrated the constructed knowledge graph to guide the response generation process. The system utilized Gemini 2.5 Flash as the primary model, while Gemini 2.5 Pro was referenced as the target model for knowledge-graph–guided reasoning and response formulation. The language model was configured with a temperature of 0.2 to produce deterministic and consistent outputs and a maximum token limit of 7,960, allowing the model to process extensive contextual information derived from the knowledge graph.

During retrieval, structured information from the knowledge graph was queried to provide relevant intervention pathways and causal chains. The system retrieved up to 5 interventions, 10 poverty-related causal chains, and 8 general effective interventions to support reasoning about the participant’s situation. These retrieved graph elements were incorporated into the model’s context to enable structured reasoning over relationships between causes, interventions, and outcomes, thereby supporting more clinically informed and explainable responses. We adopted the same configuration and hyperparameter settings across all models in the KG-based experiment to ensure a fair and consistent comparison.

\begin{figure*}[!htp]
\centering
\includegraphics[width=0.9\linewidth]{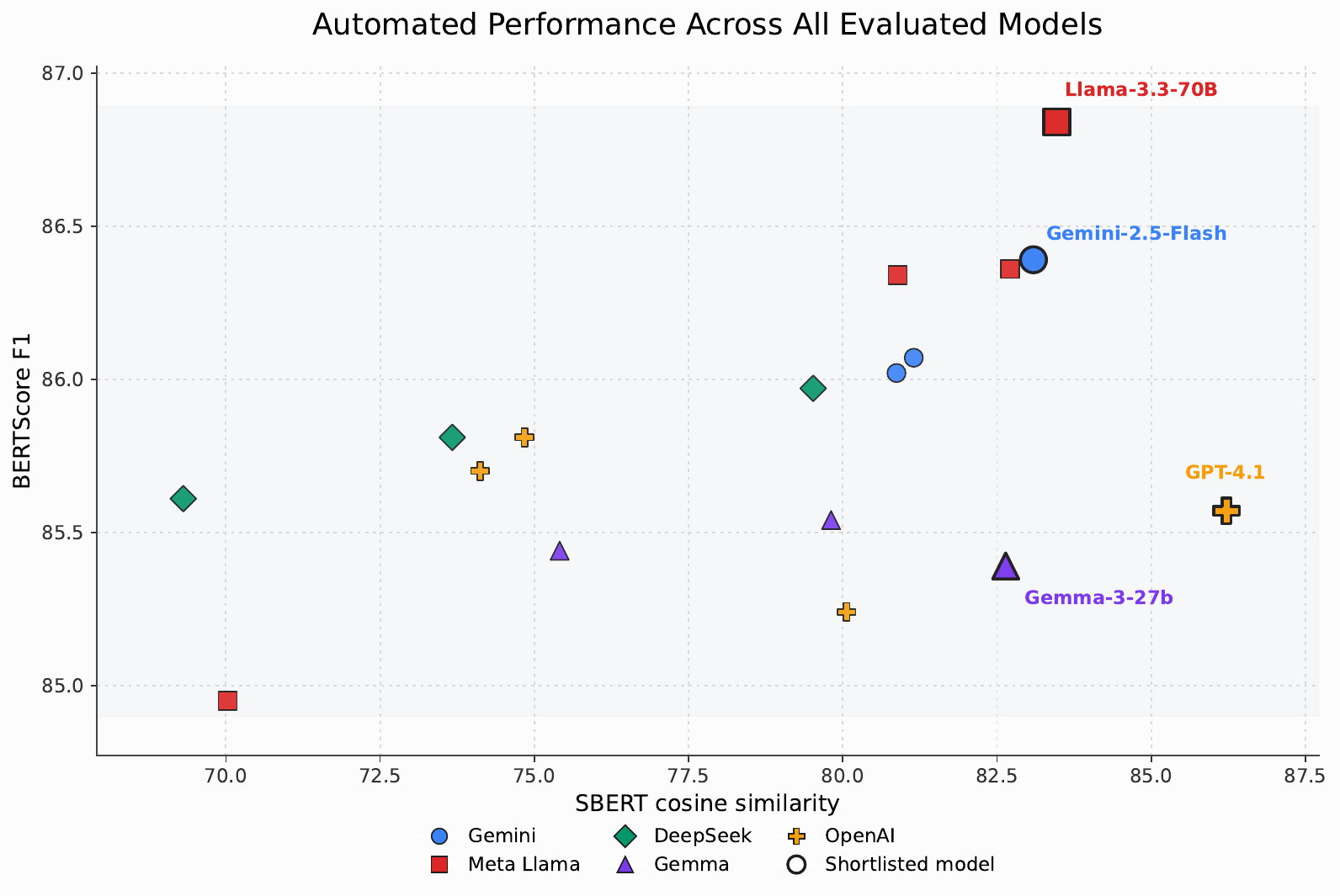}
\caption{Automated performance across all evaluated models, with SBERT cosine similarity on the horizontal axis and BERTScore F1 on the vertical axis. The plot shows that BERTScore values vary only modestly across models, whereas SBERT values exhibit greater dispersion. The four shortlisted models are labeled to indicate the candidates taken forward to human evaluation.}
\vspace{-0.4cm}
\label{fig:model-scatter}
\end{figure*}

\subsection{Evaluation Metrics}
We conducted an automatic evaluation to quantitatively assess the quality of the generated responses. We used BERTScore F1 to measure content quality, as it evaluates semantic similarity between generated responses and reference texts by comparing contextual token embeddings. This metric helps capture whether the generated content preserves the meaning and key information of the reference responses. Furthermore, we used SBERT cosine similarity to assess overall semantic similarity, measuring how closely the generated responses align with the reference responses at the sentence level. Together, these metrics provide complementary insights into both the semantic accuracy and overall similarity of the generated outputs.

In addition to automatic evaluation, we conducted a human evaluation involving PWLE, two PHRs, Psychologists, and Psychiatrists, to assess the quality and practical usefulness of the generated responses using a 5-point Likert scale (1 = best, 5 = worst). The evaluation focused on multiple dimensions relevant to counseling quality and contextual understanding. The first criterion evaluated \textit{wording}, including clarity, tone, and cultural appropriateness of the responses. The second criterion measured \textit{problem analysis accuracy}, assessing whether the system correctly identified the relevant issues and provided meaningful insights. The third criterion focused on problem-solving \textit{guidance}, examining whether the responses accurately understood the client’s situation and offered practical and relevant guidance that could be used during counseling sessions. The evaluators assessed the \textit{treatment model or intervention suggestions}, specifically the appropriateness of recommended problem-solving strategies or behavioral activation steps. Finally, \textit{environmental analysis} was evaluated to determine whether the responses considered broader contextual factors such as family dynamics, personal circumstances, economic conditions, and social environment. These evaluation dimensions allowed us to systematically measure both the clinical relevance and contextual sensitivity of the generated responses.


%% file: sections/results.tex
\section{Results and Discussion}
\label{sec:results}

\begin{figure*}[t]
\centering
\includegraphics[width=0.90\linewidth]{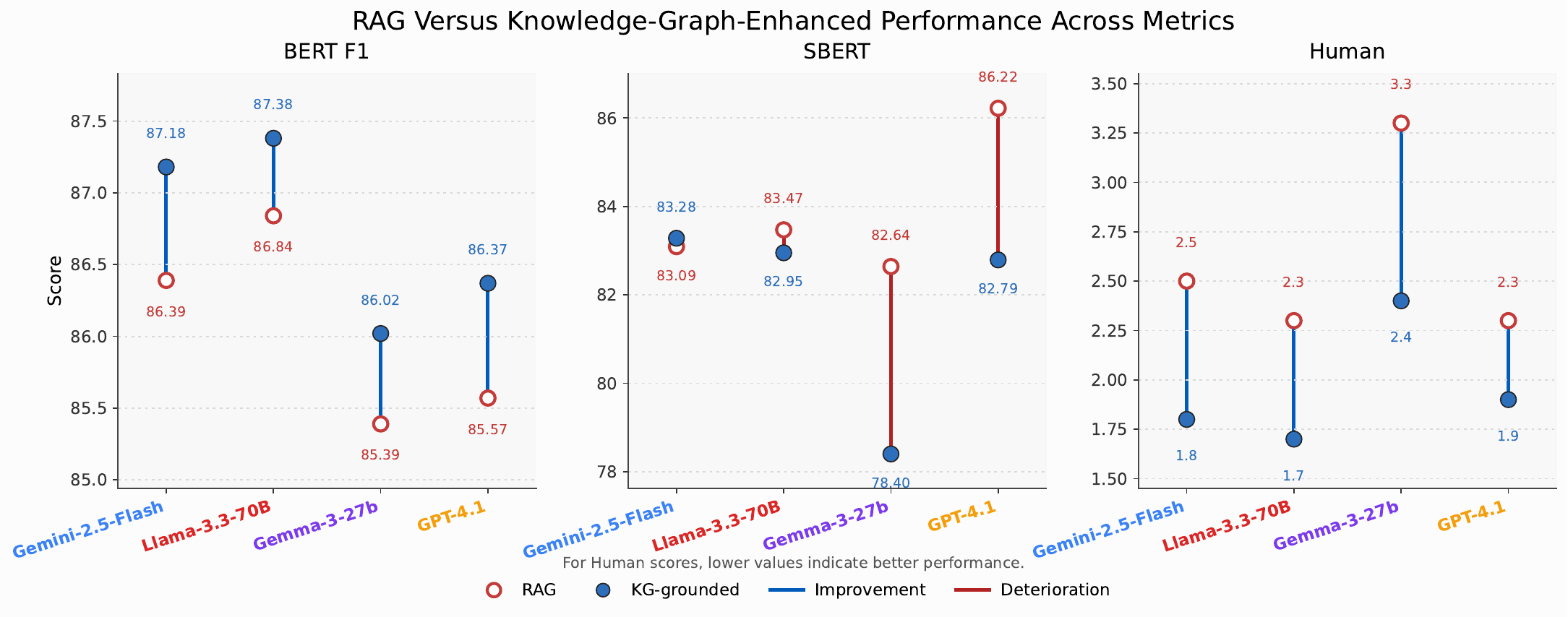}
\vspace{-0.2cm}
\caption{Comparison of RAG and knowledge-graph-grounded performance across BERTScore F1, SBERT cosine similarity, and mean human rating for the four shortlisted models. Connecting lines indicate the change from the RAG-based model to its knowledge-graph-grounded counterpart. Knowledge-graph grounding consistently improved BERTScore F1 and human ratings, while SBERT changes were mixed. 
}
\vspace{-0.4cm}
\label{fig:kg-comparison-by-metric}
\end{figure*}

We first compared seventeen candidate models from the Gemini, Llama, DeepSeek, Gemma, and OpenAI families for the RAG-based experimental setting. The automated screening results, reported in Table~\ref{tab:full-models}, show a narrow range of BERTScore F1 values and a much wider range of SBERT cosine values. As shown in Figure~\ref{fig:model-scatter}, BERTScore F1 ranged from 84.95 to 86.84, whereas SBERT cosine ranged from 69.31 to 86.22. GPT-4.1 achieved the highest SBERT cosine score, at 86.22, indicating the strongest overall structural and thematic similarity to the target counseling guide. Llama-3.3-70B achieved the highest BERTScore F1, at 86.84, indicating the strongest content alignment under that metric.

These results motivated a human comparison of four shortlisted models: Gemini-2.5-Flash, Llama-3.3-70B, GPT-4.1, and Gemma-3-27b. Table~\ref{tab:base-human-models} reports human scores at the category-level. Llama-3.3-70B and GPT-4.1 tied on the overall average, with both models scoring 2.3, but their strengths differed across categories. Llama-3.3-70B performed best on wording and problem analysis, whereas GPT-4.1 performed best on treatment and intervention quality and on environmental analysis. Gemini-2.5-Flash followed with an average of 2.5. Gemma-3-27b received the weakest human scores across all five categories, with an average of 3.3.

\begin{table}[!htp]
\centering
\footnotesize
\setlength{\tabcolsep}{2pt}
\scalebox{0.72}{
\begin{tabular}{lcccccc}
\toprule
\textbf{Model} & \textbf{Wording} & \textbf{Problem} & \textbf{Guidance} & \textbf{Treatment} & \textbf{Environ.} & \textbf{Average} \\
\midrule
Llama-3.3-70B     & 2.0 & 2.5 & 2.0 & 2.5 & 2.5 & 2.3 \\
GPT-4.1           & 2.5 & 3.0 & 2.0 & 2.0 & 2.0 & 2.3 \\
Gemini-2.5-Flash  & 2.5 & 2.5 & 2.5 & 2.5 & 2.5 & 2.5 \\
Gemma-3-27b       & 3.0 & 3.5 & 3.5 & 3.0 & 3.5 & 3.3 \\
\bottomrule
\end{tabular}
}
\caption{Human evaluation of the four shortlisted RAG-based models. Scores are mean human ratings on a five-point scale, with lower values indicating better performance. Environ.: Environmental Analysis.}
\vspace{-0.4cm}
\label{tab:base-human-models}
\end{table}


We next evaluated knowledge-graph grounding on the four shortlisted models. The knowledge graph encoded 308 nodes and 642 relationships spanning causes, effects, interventions, outcomes, categories, and counseling logic. Table~\ref{tab:kg-comparison} compares each RAG-based model with its knowledge-graph-enhanced counterpart. BERTScore F1 increased for all four models, with gains ranging from 0.54 points for Llama-3.3-70B to 0.80 points for GPT-4.1. This indicates that knowledge-graph grounding consistently improved content alignment with the target guidance.

\begin{table}[!htp]
\centering
\footnotesize

\setlength{\tabcolsep}{2pt}
\scalebox{0.73}{
\begin{tabular}{l c c c c c c c c c}
\toprule
& \multicolumn{3}{c}{\textbf{BERT F1}} & \multicolumn{3}{c}{\textbf{SBERT}} & \multicolumn{3}{c}{\textbf{Human}} \\
\cmidrule(lr){2-4}\cmidrule(lr){5-7}\cmidrule(lr){8-10}
\textbf{Model}
& \textbf{RAG} & \textbf{KG} & \textbf{$\Delta$}
& \textbf{RAG} & \textbf{KG} & \textbf{$\Delta$}
& \textbf{RAG} & \textbf{KG} & \textbf{$\Delta$} \\
\midrule
Gemini-2.5-Flash
& 86.39 & 87.18 & \textcolor{BetterBlue}{$\uparrow\,0.79$}
& 83.09 & 83.28 & \textcolor{BetterBlue}{$\uparrow\,0.19$}
& 2.50 & 1.80 & \textcolor{BetterBlue}{$\downarrow\,0.70$} \\

Llama-3.3-70B
& 86.84 & 87.38 & \textcolor{BetterBlue}{$\uparrow\,0.54$}
& 83.47 & 82.95 & \textcolor{WorseRed}{$\downarrow\,0.52$}
& 2.30 & 1.70 & \textcolor{BetterBlue}{$\downarrow\,0.60$} \\

Gemma-3-27b
& 85.39 & 86.02 & \textcolor{BetterBlue}{$\uparrow\,0.63$}
& 82.64 & 78.40 & \textcolor{WorseRed}{$\downarrow\,4.24$}
& 3.30 & 2.40 & \textcolor{BetterBlue}{$\downarrow\,0.90$} \\

GPT-4.1
& 85.57 & 86.37 & \textcolor{BetterBlue}{$\uparrow\,0.80$}
& 86.22 & 82.79 & \textcolor{WorseRed}{$\downarrow\,3.43$}
& 2.30 & 1.90 & \textcolor{BetterBlue}{$\downarrow\,0.40$} \\
\bottomrule
\end{tabular}
}

\caption{Comparison of RAG-based and knowledge-graph-enhanced models. BERT F1 denotes BERTScore F1, SBERT denotes SBERT cosine similarity, and Human denotes the mean human rating on a five-point scale, with lower values indicating better performance. In the $\Delta$ columns, \textcolor{BetterBlue}{blue} indicates improvement and \textcolor{WorseRed}{red} indicates deterioration relative to the RAG model.}
\vspace{-0.4cm}
\label{tab:kg-comparison}
\end{table}

The SBERT results show a less consistent pattern. Gemini-2.5-Flash improved slightly after knowledge-graph grounding. In contrast, Llama-3.3-70B, Gemma-3-27b, and GPT-4.1 showed lower SBERT cosine scores after grounding, with the largest decrease observed for Gemma-3-27b. These mixed changes suggest that knowledge-graph grounding altered the structure or phrasing of the generated responses in ways that did not always increase sentence-level similarity to the reference text, even when the content became more closely aligned.

The human evaluation provides a clearer picture for the present application. All four models improved after knowledge-graph grounding. The average human score decreased by 0.6 points for Llama-3.3-70B, 0.4 points for GPT-4.1, 0.7 points for Gemini-2.5-Flash, and 0.9 points for Gemma-3-27b. After grounding, Llama-3.3-70B achieved the strongest mean human score, at 1.7. Gemini-2.5-Flash followed closely at 1.8, and GPT-4.1 followed at 1.9. Although Gemma-3-27b remained the weakest of the four models overall, it showed the largest gain in average human score. Table~\ref{tab:kg-human-categories} reports the underlying category-level ratings before and after grounding across \textit{wording, problem analysis, guidance, treatment and intervention quality,} and \textit{environmental analysis}.


Figure~\ref{fig:kg-comparison-by-metric} shows that model choice affects performance, but no single model consistently outperforms others across all evaluation settings. Automated metrics favored different models (GPT-4.1 for overall similarity and Llama-3.3-70B for BERTScore F1) while human evaluation rated them similarly overall, highlighting differences mainly in specific categories. This indicates that automated evaluation alone is insufficient, as responses may match reference texts without being clear, appropriate, or practically useful for counseling.

The findings also demonstrate that expert-curated knowledge and knowledge-graph grounding had a stronger impact on performance than solely model selection. Incorporating structured, clinically reviewed knowledge improved results across all tested models, particularly in real-world use. However, environmental and contextual analysis remained weaker than other aspects, such as wording, guidance, and problem analysis. Future work should therefore focus on better representing family, personal, and economic contexts and improving support for complex interpersonal situations in counseling.

%% file: sections/conclusion.tex
\section{Conclusion}
\label{sec:conclusion}

In this work, we propose RAG- and KG-based approaches for generating counseling support responses by integrating structured domain knowledge with large language models in Bangladesh context. We first curated and manually annotated the narratives of the counseling sessions with a multidisciplinary team 
to ensure compliance with privacy, narrative coherence, and clinical validity. Using validated data, we constructed a domain-specific knowledge graph to represent causal relationships among stressors, interventions, and effects. We then compared two experimental settings, RAG- and KG–based experiments, across multiple language models to examine whether incorporating culturally-grounded knowledge improves counseling quality. Our results show that incorporating structured, clinically curated knowledge from para-counselors, PHRs, PWLEs, psychologists, and psychiatrists consistently improves the quality and practical relevance of generated counseling responses more than model selection alone. Human evaluation 
highlights that effective counseling support depends not only on semantic similarity to reference responses but also on clear wording, accurate problem interpretation, appropriate guidance, and contextual awareness. However, environmental and broader contextual analysis remain a challenging aspect for current systems.

Future work will focus on expanding the knowledge graph with richer contextual factors, improving the representation of family, personal, and socioeconomic contexts, and developing more robust evaluation protocols for counseling-oriented language generation systems.

%% file: sections/appendix.tex
\section{Dataset}

\subsection{Guideline for Conducting Sessions}
\label{appn:guideline_session}

\paragraph{Session 1} The goal of this session is to establish understanding with participants and build trust in the para-counselors. Participants are reassured that the para-counselors are there to support them and that all shared information will remain confidential. This environment is intended to help participants feel comfortable discussing their personal concerns and emotional experiences openly and without hesitation. Para-counselors were additionally instructed to respond adaptively to participants’ circumstances. For instance, if a participant disclosed the death of a relative, para-counselors were expected to provide appropriate emotional support and empathy. This session primarily focuses on identifying the key sources of the participant’s stress and emotional condition, which can subsequently inform targeted and actionable counseling interventions. The goal of the first session is two-fold. First is to give an opportunity to both the participant and the para-counselor to understand each other. Second, help the participant build trust in the para-counselor by ensuring they are available when needed and protecting the confidentiality of information shared by participants, among other measures. 

\paragraph{Session 2} The objective of this session is to transition from problem identification to a solution-oriented discussion. Para-counselors work with participants to explore practical strategies for addressing their challenges. For example, for financial concerns, counselors can discuss budgeting or savings practices, while for interpersonal or domestic conflicts, they can explore behavioral adjustments, anger management strategies, or appropriate support-seeking actions. This session also follows a structured guideline that prioritizes which issues should be addressed first, ensuring that the most urgent or impactful concerns are discussed before others. Moreover, participants are guided to select a small set of activities that may improve mood or daily functioning. Para-counselors also introduce an activity log to track behaviors and emotions and schedule the next appointment.
\paragraph{Session 3-5} These sessions focus on reviewing the participant’s activity log and discussing how the selected activities have influenced mood and daily functioning. Para-counselors ask participants about emotional changes, discuss which activities were helpful or challenging, and revisit previously identified problems. Based on the participant’s progress, new activities aligned with the goals established in Session 1 may be added.

\paragraph{Session 6} The final session summarizes the skills and strategies learned throughout the counseling process. Para-counselors review the participant’s goals, discuss progress, and help the participant develop a practical plan for responding to future challenges. Participants are also encouraged to create a ``toolbox" of helpful coping strategies that they can use independently.

\subsection{Annotation Guideline}
\label{sec:annotation_guideline}
Our annotation guideline for reviewing and annotating session narratives in two stages: \textit{i)} Appropriateness and Data Privacy Review and \textit{ii)} Expert Clinical Review. Annotators should carefully evaluate each case to ensure accuracy, completeness, privacy protection, and clinical soundness.

\subsubsection{Appropriateness and Data Privacy Review}
In this step, the annotators evaluate each session narrative for privacy compliance, clarity, and structural validity. Each case should be marked as Appropriate or Not Appropriate, with suggested corrections provided where necessary. Annotators must also correct spelling and grammar errors and remove any identifying details. The annotators also performed data anonymization and de-identification, along with session narrative appropriateness, using the following criteria:

\paragraph{Data anonymization and De-identification} Annotators must remove all direct identifiers, including names, addresses, phone numbers, workplaces, schools, and specific community names. Replace these with generalized descriptors (e.g., `the client', `a family member', `a local workplace'). Specific calendar dates should be generalized using relative references such as `Session 1', `Week 2', or `7 days after the first session' to preserve temporal structure without revealing identifiable information.

\paragraph{Session Narrative Appropriateness} Annotators must verify whether demographics, contextual factors, presenting problems, stressors, and protective factors are correctly identified and categorized. Moreover, annotators should assess whether important causes or protective factors have been overlooked or misclassified. The narrative must clearly reflect a coherent causal chain: stressors or causes $\rightarrow$ psychological/behavioral effects $\rightarrow$ intervention applied $\rightarrow$ resulting outcomes. If any component is missing or logically inconsistent, it should be flagged for revision.

\subsubsection{Expert Clinical Review}
In this annotation step, a qualified psychologist evaluates the clinical integrity and completeness of the session narrative. Cases that fail to clearly identify a causal chain and intervention outcome should be marked as \textit{Incomplete}. The psychologist should determine whether demographics, contextual variables, stressors, protective factors, and presenting problems are clinically appropriate and accurately interpreted. Particular attention should be paid to potential discrepancies in the classification of protective factors. The session narrative must provide sufficient information to establish a meaningful cause and effect relationship between identified stressors, intervention strategies, and observed outcomes.

In Addition, the psychologist should evaluate whether the selected intervention type is clinically appropriate given the client’s demographics and presenting concerns. Finally, the psychologists must assess whether any point in the causal chain requires an emergency guardrail (e.g., crisis referral, notification of family members, psychiatric evaluation, or hospitalization). Missing but necessary safeguards should be explicitly identified.

\subsection{Data Analysis}
\label{appn:data_analysis}
We conducted descriptive analysis to understand the demographic characteristics present in the annotated dataset of 69 cases. The dataset is predominantly composed of female participants (65), with 4 male participants. Regarding marital status, 51 participants were married, 3 were widowed, 1 was divorced, 2 were unmarried, and 12 cases did not specify marital status. Table~\ref{tab:literacy_level} presents the literacy levels that are generally considered a low level of educational attainment. Among the 40 participants with very limited literacy, 33 are able to sign only their names, and 7 are unable to sign their names at all. Among the remaining participants, 13 had primary education (Grades 1–5), 12 had education up to Grades 6–10, and 4 had completed secondary or higher secondary education (SSC/HSC).

\begin{table}[!ht]
\centering
\scalebox{0.69}{
\begin{tabular}{cc|ccc}
\toprule
\multicolumn{2}{c}{\textbf{Illiterate}}& \multicolumn{3}{c}{Literate} \\ \hline
Signature & No Signature & Grade 1-5 & Grade 6-10 & SSC/HSC \\
\hline
33 & 7 & 13 & 12 & 4 \\
\bottomrule
\end{tabular}
}
\caption{Literacy level of participants in the dataset. SSC: Secondary School Certificate, HSC: Higher Secondary Certificate.}
\vspace{-0.4cm}
\label{tab:literacy_level}
\end{table}

We also performed an analysis of the occupations of the participants. Table~\ref{tab:occupations} presents the detailed occupation list, where many participants were engaged in informal or unstable employment. The most common categories include unemployed who are not interested in finding work (NIFW), housemaids, unemployed who are actively looking for work, and small business workers. Other occupations included waste collectors, any job that pays monthly, and a small number of street people, rickshaw drivers, and daily laborers. Overall, the analysis highlights that the dataset largely represents women from low socioeconomic and educational backgrounds.

\begin{table}[!htp]
\centering
\begin{tabular}{lr}
\toprule
\textbf{Occupation} & \textbf{Count} \\ \midrule
Unemployed (NIFW) & 17 \\
Waste Collector & 5 \\
Any Job & 4 \\
Unemployed (LW) & 15 \\
Housemaid & 15 \\
Small Business & 10 \\
Rickshaw Driver & 1 \\
Daily Labor & 1 \\
Street People & 1 \\
\bottomrule
\end{tabular}
\caption{Occupation of the participants in the dataset. NIFW: not interested in finding work, LW: looking for work.}
\vspace{-0.4cm}
\label{tab:occupations}
\end{table}

\begin{table}[!htp]
\centering
\footnotesize

\setlength{\tabcolsep}{4pt}

\begin{tabular}{llcc}
\toprule
\textbf{Model} & \textbf{Family} & \textbf{BERT F1} & \textbf{SBERT} \\
\midrule
gemini-2.5-pro     & \multirow{3}{*}{Gemini}     & 86.07 & 81.15 \\
gemini-2.5-flash   &     & 86.39 & 83.09 \\
gemini-2.0-flash   &    & 86.02 & 80.87 \\ \midrule
Llama-4-Maverick   & \multirow{4}{*}{Llama} & 86.36 & 82.71 \\
Llama-3.3-70B      & & \textbf{86.84} & 83.47 \\
Llama-3.1-70B      & & 86.34 & 80.89 \\
Llama-3-8B         & & 84.95 & 70.03 \\ \midrule
DeepSeek-V3.2      & \multirow{3}{*}{DeepSeek}   & 85.97 & 79.52 \\
DeepSeek-V3.1      &   & 85.61 & 69.31 \\
DeepSeek-R1        &   & 85.81 & 73.67 \\ \midrule
gemma-3-27b        & \multirow{3}{*}{Gemma}     & 85.39 & 82.64 \\
gemma-3-12b        &     & 85.44 & 75.41 \\
gemma-3-4b         &     & 85.54 & 79.81 \\ \midrule
GPT-5.1            & \multirow{4}{*}{OpenAI}     & 85.81 & 74.84 \\
GPT-4.1            &      & 85.57 & \textbf{86.22} \\
GPT-4o             &     & 85.70 & 74.12 \\
gpt-oss-20b        &     & 85.24 & 80.06 \\
\bottomrule
\end{tabular}

\caption{Automated evaluation for all seventeen models. BERT F1 denotes BERTScore F1, and SBERT denotes SBERT cosine similarity. \textbf{Bold} indicates best performance.}
\label{tab:full-models}

\end{table}

\section{Results}
\label{appendix:results}

Table~\ref{tab:full-models} reproduces the automated scores for all seventeen evaluated models. These values motivated the shortlist used for human evaluation in the main text. Moreover, Table~\ref{tab:kg-human-categories} shows category-level human scores before and after knowledge-graph grounding for the four shortlisted models.

\begin{figure*}[t]
\centering
\includegraphics[width=0.75\linewidth]{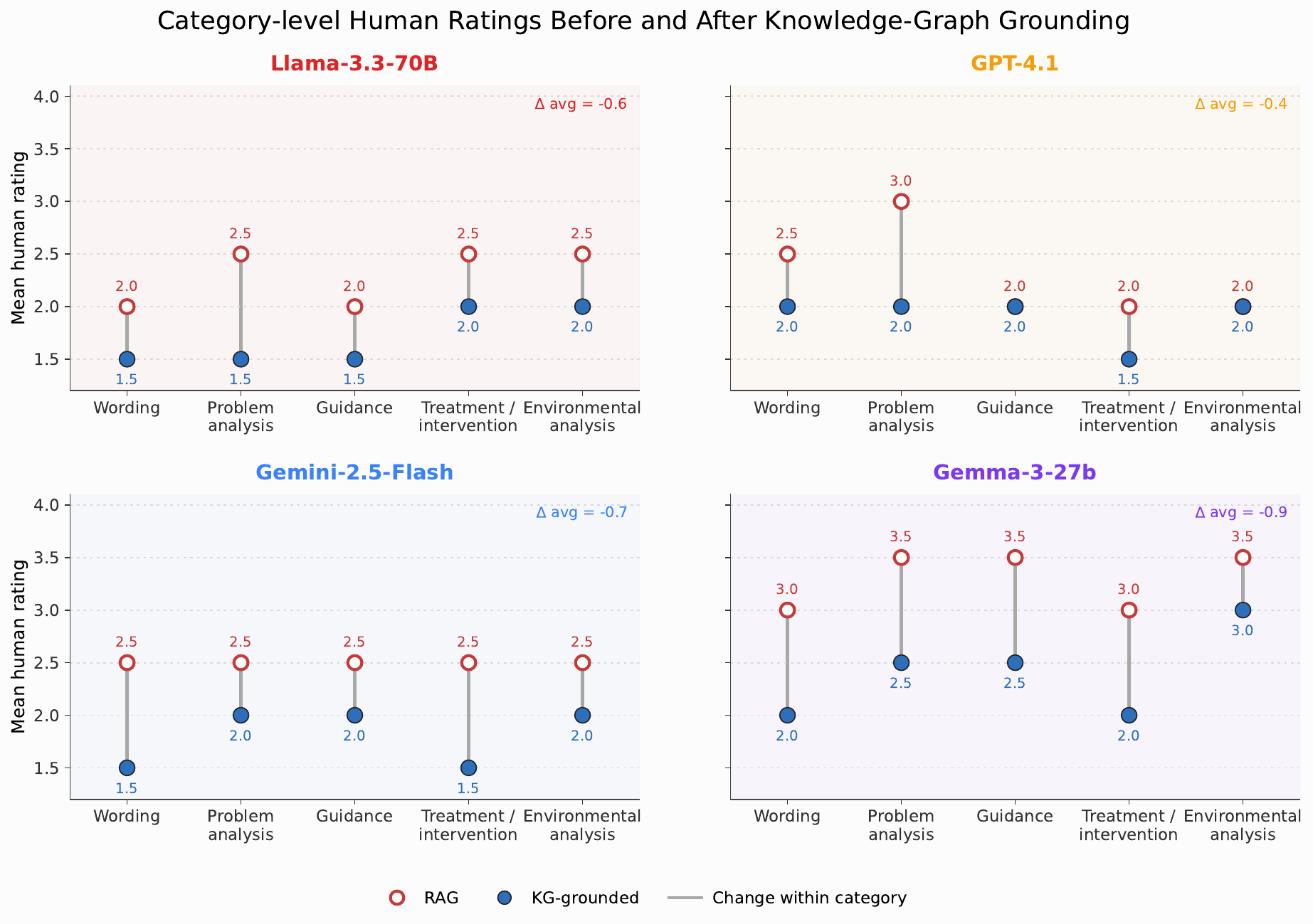}
\caption{Category-level human ratings before and after knowledge-graph grounding for the four shortlisted models. Lower values indicate better performance. The figure shows paired category-wise ratings for each model, highlighting consistent decreases in average human scores after grounding, with variation in the magnitude of change across categories and models.}
\vspace{-0.4cm}
\label{fig:kg-items}
\end{figure*}

\begin{table*}[]
\centering
\footnotesize

\setlength{\tabcolsep}{3.5pt}
\renewcommand{\arraystretch}{1.1}
\begin{tabular}{lc c c c c c c c c c c c}
\toprule
& \multicolumn{3}{c}{\textbf{Llama-3.3-70B}} & \multicolumn{3}{c}{\textbf{GPT-4.1}} & \multicolumn{3}{c}{\textbf{Gemini-2.5-Flash}} & \multicolumn{3}{c}{\textbf{Gemma-3-27b}} \\
\cmidrule(lr){2-4}\cmidrule(lr){5-7}\cmidrule(lr){8-10}\cmidrule(lr){11-13}
\textbf{Category} & \textbf{RAG} & \textbf{KG} & \textbf{$\Delta$} & \textbf{RAG} & \textbf{KG} & \textbf{$\Delta$} & \textbf{RAG} & \textbf{KG} & \textbf{$\Delta$} & \textbf{RAG} & \textbf{KG} & \textbf{$\Delta$} \\
\midrule
Wording
& 2.0 & 1.5 & \textcolor{BetterBlue}{$\downarrow\,0.5$}
& 2.5 & 2.0 & \textcolor{BetterBlue}{$\downarrow\,0.5$}
& 2.5 & 1.5 & \textcolor{BetterBlue}{$\downarrow\,1.0$}
& 3.0 & 2.0 & \textcolor{BetterBlue}{$\downarrow\,1.0$} \\

Problem analysis
& 2.5 & 1.5 & \textcolor{BetterBlue}{$\downarrow\,1.0$}
& 3.0 & 2.0 & \textcolor{BetterBlue}{$\downarrow\,1.0$}
& 2.5 & 2.0 & \textcolor{BetterBlue}{$\downarrow\,0.5$}
& 3.5 & 2.5 & \textcolor{BetterBlue}{$\downarrow\,1.0$} \\

Guidance
& 2.0 & 1.5 & \textcolor{BetterBlue}{$\downarrow\,0.5$}
& 2.0 & 2.0 & $0.0$
& 2.5 & 2.0 & \textcolor{BetterBlue}{$\downarrow\,0.5$}
& 3.5 & 2.5 & \textcolor{BetterBlue}{$\downarrow\,1.0$} \\

Treatment / intervention
& 2.5 & 2.0 & \textcolor{BetterBlue}{$\downarrow\,0.5$}
& 2.0 & 1.5 & \textcolor{BetterBlue}{$\downarrow\,0.5$}
& 2.5 & 1.5 & \textcolor{BetterBlue}{$\downarrow\,1.0$}
& 3.0 & 2.0 & \textcolor{BetterBlue}{$\downarrow\,1.0$} \\

Environmental analysis
& 2.5 & 2.0 & \textcolor{BetterBlue}{$\downarrow\,0.5$}
& 2.0 & 2.0 & $0.0$
& 2.5 & 2.0 & \textcolor{BetterBlue}{$\downarrow\,0.5$}
& 3.5 & 3.0 & \textcolor{BetterBlue}{$\downarrow\,0.5$} \\

\textbf{Average}
& \textbf{2.3} & \textbf{1.7} & \textbf{\textcolor{BetterBlue}{$\downarrow\,0.6$}}
& \textbf{2.3} & \textbf{1.9} & \textbf{\textcolor{BetterBlue}{$\downarrow\,0.4$}}
& \textbf{2.5} & \textbf{1.8} & \textbf{\textcolor{BetterBlue}{$\downarrow\,0.7$}}
& \textbf{3.3} & \textbf{2.4} & \textbf{\textcolor{BetterBlue}{$\downarrow\,0.9$}} \\
\bottomrule
\end{tabular}

\caption{Category-level human scores before and after knowledge-graph grounding for the four shortlisted models. Scores are mean human ratings on a five-point scale, with lower values indicating better performance. In the $\Delta$ columns, \textcolor{BetterBlue}{blue} indicates improvement relative to the RAG model.}
\label{tab:kg-human-categories}
\end{table*}